\documentclass[sigconf,anonymous=false,screen,nonacm]{acmart}
\usepackage{amsmath}
\usepackage{graphicx}
\usepackage{ulem}
 \usepackage{hyperref}
 \hypersetup{
     colorlinks=true,
     linkcolor=blue,
     filecolor=blue,
     citecolor=black,      
     urlcolor=cyan,
}
\usepackage{subcaption}
\usepackage{tikz}
\usepackage{diagbox}
\usepackage{tikzsymbols}
            
\definecolor{dollarbill}{rgb}{0.52, 0.73, 0.4}
\newcommand{\good}[1]{\textcolor{dollarbill}{{\bf #1}}}
\definecolor{gold(metallic)}{rgb}{0.83, 0.69, 0.22}
\newcommand{\neutral}[1]{\textcolor{gold(metallic)}{{\bf #1}}}
\definecolor{carmine}{rgb}{0.59, 0.0, 0.09}

\newcommand{\nllh}{ {NCE} }

\newcommand*\circled[1]{\tikz[baseline=(char.base)]{
            \node[shape=circle,draw,inner sep=1.5pt] (char) {#1};}}

\makeatletter
\def\@copyrightspace{\relax}
\makeatother

\begin{document}

\title{Lessons from the AdKDD'21 Privacy-Preserving ML Challenge}

\author{Eustache Diemert}
\affiliation{%
\institution{Criteo AI Lab}
}

\author{Romain Fabre}
\affiliation{%
\institution{Criteo AI Lab}
}

\author{Alexandre Gilotte}
\affiliation{%
\institution{Criteo AI Lab}
}

\author{Fei Jia }
\affiliation{%
\institution{Meta Platforms, Inc.}
}

\author{Basile Leparmentier}
\affiliation{%
\institution{Criteo AI Lab}
}

\author{Jérémie Mary}
\affiliation{%
  \institution{Criteo AI Lab}
}

\author{Zhonghua Qu}
\affiliation{%
  \institution{Meta Platforms, Inc.}
}

\author{Ugo Tanielian}
\affiliation{%
  \institution{Criteo AI Lab}
}

\author{Hui Yang}
\affiliation{%
  \institution{Meta Platforms, Inc.}
}
\renewcommand{\shortauthors}{E. Diemert \& al.}

\begin{abstract}
Designing data sharing mechanisms providing performance and strong privacy guarantees is a hot topic for the Online Advertising industry.
Namely, a prominent proposal discussed under the Improving Web Advertising Business Group at W3C only allows sharing advertising signals through aggregated, differentially private reports of past displays.
To study this proposal extensively, an open Privacy-Preserving Machine Learning Challenge took place at AdKDD'21, a premier workshop on Advertising Science with data provided by advertising company Criteo.
In this paper, we describe the challenge tasks, the structure of the available datasets, report the challenge results, and enable its full reproducibility.
A key finding is that learning models on large, aggregated data in the presence of a small set of unaggregated data points can be surprisingly efficient and cheap.
We also run additional experiments to observe the sensitivity of winning methods to different parameters such as privacy budget or quantity of available privileged side information. We conclude that the industry needs either alternate designs for private data sharing or a breakthrough in learning with aggregated data only to keep ad relevance at a reasonable level.
\end{abstract}



\maketitle

\section{Introduction}

\paragraph{Motivation}
Motivated by changes in legislation and users pressure, all major browser vendors are restricting the possibilities to track the behavior of users or have plans to do it. This new operational constraint initiated a deep mutation of the Online Advertising ecosystem, which remains central to the funding of a large part of the open internet. 
Current discussions involving advertisers, publishers, and technologists at W3C focus on methods allowing to personalize advertising - thus preserving ad relevance and efficiency - while providing strong privacy guarantees to users. One major proposal, supported by Google in the Chrome Privacy Sandbox, is the Measurement API \citep{aggreportapi} which allows advertisers to access limited data from user browsers in the form of aggregated differential private reports.

To allow the development of an Open Internet outside of the walled gardens, it is crucial to offer a competitive economical model. As users are demanding for personalization, we need to understand the impact of the privacy-preserving tools on the ability to use Machine Learning. As this will shape the future of the Internet we think these should be discussed by the community.
To the extent of our knowledge, this was the first realistic benchmark for machine learning operating under future privacy constraints in the industry. However, it is still unclear whether these noisy aggregated data still offer the possibility to learn relevant prediction models for ad placement. To assess and discuss this point openly, \href{http://www.criteo.com}{Criteo} - a leader in performance advertising - donated a new dataset to organize a challenge at \href{https://www.adkdd.org}{AdKDD'21} on this topic.

\paragraph{Overview of Learning with Aggregated, Differentially Private Data}

\begin{figure}[t]
    \centering
    \includegraphics[width=1.0\linewidth]{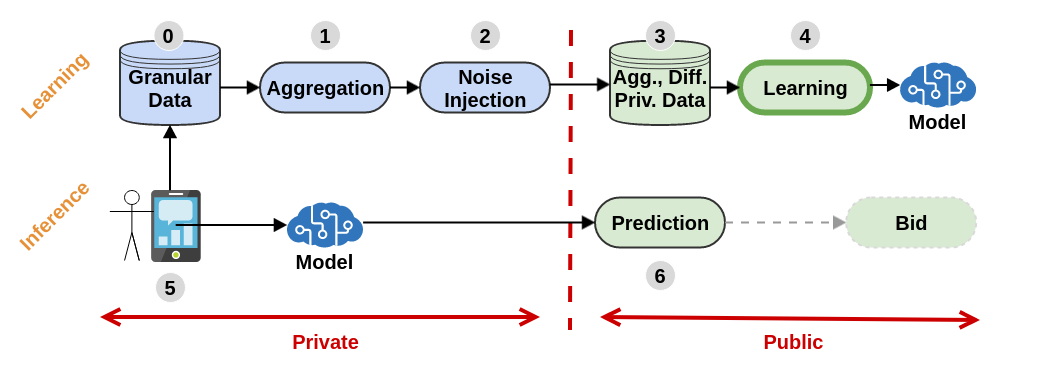}
    \caption{Overview of Machine Learning using Aggregated, Differentially Private Data}
    \label{fig:overview}
\end{figure}

In the future, a pipeline (summarized in Fig. \ref{fig:overview}) conforming to the Measurement API proposal might be organized as follows. To begin with,  users generate private, granular data \circled{0} for which you can see an example in Table \ref{tab:sample_granular_data}(a). This data is then aggregated by the internet browser \circled{1}; this corresponds to samples in Table \ref{tab:sample_granular_data}(b). Furthermore, noise is injected \circled{2} according to a given privacy budget, and this produces an aggregated, differentially private report \circled{2} as depicted in Table \ref{tab:sample_granular_data}(c). The dataset is then available to demand-side platforms (i.e. advertisers and agencies), which can learn prediction models \circled{4} for marketing outcomes such as clicks and sales. Such models are finally deployed inside a user browser \circled{5} and called whenever an ad slot becomes available to make predictions and place a bid in the online advertising auction \circled{6}.


\begin{table*}
\begin{subtable}[t]{.3\linewidth}
    \centering
    \begin{tabular}{l|l|l|l|l}
         Feat.1 & Feat.2 & \ldots & Feat.N & Click\\
         \hline
         3 & A & \ldots & aef & 0\\
         3 & A & \ldots & z3f & 1\\
         7 & B & \ldots & 4eh & 0\\
         8 & B & \ldots & aef & 1\\
         8 & B & \ldots & 66e & 0\\
         \ldots & \ldots & \ldots & \ldots & \ldots\\
    \end{tabular}
\caption{Sample Granular Data}
\end{subtable}%
\begin{subtable}[t]{.345\linewidth}
    \centering
    \begin{tabular}{l|l|r|r}
         Feat.1 & Feat.2 & Count & Sum(Click)\\
         \hline
         3 & A & 1,000 & 123\\
         3 & B & 3,000 & 23\\
         8 & B & 225 & 111\\
         7 & B & 22,500 & 1,711\\
         \ldots & \ldots & \ldots & \ldots\\
         \ldots & \ldots & \ldots & \ldots\\
    \end{tabular}
\caption{Sample Aggregated Data}
\end{subtable}%
\ 
\begin{subtable}[t]{.345\linewidth}
    \centering
    \begin{tabular}{l|l|r|r}
         Feat.1 & Feat.2 & Count & Sum(Click)\\
         \hline
         3 & A & 1,118.3 & 135.7\\
         3 & B & 3,411.1 & -2.5\\
         8 & B & 137.9 & 99.5\\
         7 & B & 22,105.8 & 1,731.6\\
         \ldots & \ldots & \ldots & \ldots\\
         \ldots & \ldots & \ldots & \ldots\\
    \end{tabular}
\caption{Sample Aggregated, Noisy Data}
\end{subtable}%
\caption{Sample Data for Learning using Aggregated, Differentially Private mechanisms
\label{tab:sample_granular_data}
}
\end{table*}

\paragraph{Challenge objective} The AdKDD'21 competition targeted the task of learning models that predict individual, granular outcomes from aggregated, noisy data \circled{4}. To that end, competitors had access to a large pre-aggregated dataset \circled{3} alongside with a tiny set of granular data. Submissions were evaluated on the prediction quality \circled{6}. 
The opportunities and difficulties to design alternative aggregation strategies are left as future works and can be studied by interested parties thanks to the public release of the complete granular data with this paper. To summarize, the contributions of the present paper are:
\begin{itemize}
    \item We release the complete dataset and evaluation code, enabling full reproducibility of the results.
    \item We summarize and analyze the winning methods, providing robustness and ablation studies.
    \item We demonstrate the key importance of the presence of a small granular dataset to fully exploit the aggregated data, opening new research directions.  
    \item We provide guidance on which methods perform best as a function of available data. 
\end{itemize}

\section{Related Works}
\label{sec:related_works}

\paragraph{Related Browser Privacy Proposals}
Major browser vendors are participating in the process of improving privacy by submitting design proposals to the Improving Web Advertising Business Group at W3C. 
The leading proposal is the Chrome Privacy Sandbox; in this work, we study how to learn a model under constraints inspired by the surrounding discussions.
In particular, the FLEDGE proposal describes a mechanism to do privacy-guaranteed targeted advertising.
In this proposal, the advertisers would upload their ML models to the user browser or a gatekeeper, where it could access to private granular data at inference time, while learning has to be done from aggregated observations.

Apple Safari implemented Private Click Measurement \cite{pcm} to obfuscate user identifiers when reporting click and conversion events by limiting the entropy of identifiers. It is still unclear how it would affect data that is usable to learn models.
 Microsoft proposed Parakeet with the MaskedLARK reporting design \cite{maskedLARK} to address challenges of model learning under privacy constraints. Similarly to the Privacy Sandbox, it describes aggregated, differentially private reports. It also mentions K-anonimized data \cite{samarati1998protecting}, a technique akin to principled bucketing of feature values. Finally MaskedLARK proposes to perform more complex ``masked'' gradient operations on browser side, a technique allowing to implement a form of centralized Federated Learning \cite{konevcny2015federated}.

All in all we observe that in the most advanced design proposals, data will be accessible in the form of aggregated, differentially private reports which validates the relevance of the AdKDD challenge.

\paragraph{Differential Privacy and Learning} 
The concept of Differential Privacy \cite{dwork2006our} embodies an intuitive notion of privacy with statistical guarantees. An $\epsilon$-differentially private process guarantees that adding a record to a given database does not increase the probability of identifying the record by more than $\exp(\epsilon)$. Consequently, this is now used by companies and governmental agencies in the design of the treatment of sensitive information \cite{diffpriv_census, blogdefontaines}. In this challenge, we used a variant named $(\epsilon, \delta)$-differential privacy, which may be enforced by the addition of Gaussian noise to the aggregated data.

Machine learning algorithms providing differential privacy guarantees have been proposed by introducing result perturbation, objective perturbation \cite{Chaudhuri2011} or noisy iterative optimization methods. For instance, stochastic methods adding noise to training batches have shown interesting performances when applied to deep learning models \cite{Abadi2016, Papernot2017}. Unfortunately, these methods rely on the access to individual  - possibly noisy - records or the computation of complex functions like gradients on the browser side that are not possible according to the Measurement API proposal.

\paragraph{Learning from aggregated data}
Learning to predict individual-level outcomes from aggregated data is known historically as the ecological inference problem \cite{king2004ecological}. There exist different meanings of "learning from aggregated data" as different settings perform aggregation at different levels: label similarities with complete access to features \cite{zhang2020learning}, aggregated labels with access to individual features \cite{bhowmik2019estimagg} and  aggregated labels with aggregated features \cite{bhowmik2016sparse, gilotte2021learning}. The Measurement API case is of the latter nature which is the most challenging as both features and labels are considered as sensitive and thus are only available through an aggregation. 
Yet the scaling of the existing methods to the industrial level in terms of number of features and examples is still unclear. Also, differential privacy should be added on top of these methods. 

\paragraph{Ad Click Prediction Competitions}
Ad click prediction is a well-studied problem \cite{mcmahan2013ad}, at least in tradition,al settings with access to individual records. As the problem is central to a multi-billion dollar industry, companies have supported the organization of challenges to advance the state of the art, the most prominent being the iPinYou bidding competition \cite{liao2014ipinyou}, Criteo Display Advertising Challenge \cite{criteo_challenge}, Avazu CTR Prediction Contest \cite{avazu}, Outbrain Click Prediction Challenge \cite{outbrain}, and Avito Contextual Ad Clicks \cite{avito}. Even though the exact features, data size and marketing outcomes differ between these competitions the setup is standard and does not encompass the challenges of learning models from aggregated, noisy data. Yet, at Recsys'21 was introduced for the first time by \cite{belli20212021} a challenge with a fairness constraint: learning to rank tweets, if a user deleted their content, the same content would be promptly removed from the dataset and candidates had to keep the data up-to-date.

\section{Challenge Setup}
\label{sec:challenge_setup}

We now formalize the task and the evaluation metrics of the challenge, highlighting  how it relates to the current proposal for the measurement API at W3C \cite{measurementapi}.

\subsection{Introduction to the dataset}
The challenge setup is based on the available information as of May 10th, 2021, as discussed in this \href{https://github.com/WICG/conversion-measurement-api/issues/137}{github issue}. The aim was to produce the most straightforward, real and tractable experimental setup. 

The dataset provided in this challenge originally comes from a granular database with a size of about 92 million displays. This is reasonably representative of a typical advertising dataset while being easy to work with with a desktop computer. 
This dataset is composed of 19 distinct features, that were selected among the most informative of a production, performance-focused ad placement system. 
The data contains both click and sales labels with a positive rate of 10\% for clicks, and about .5\% for sales; the latter presents a more complex and noisy learning problem. 

Those data have been randomly split into three different datasets: 
\begin{enumerate}
    \item the main dataset $\mathcal{D}_{raw}$ of 88M samples, which was not released in the competition, and from which the noisy aggregated dataset $\mathcal{D}_{agg}$ was computed,
    \item a tiny set (1,000x smaller) $\mathcal{D}_{train}$ made of granular, noiseless, and labeled data,
    \item a test set $\mathcal{D}_{test}$ of about 1M granular, noiseless, but unlabeled data used for evaluation purposes.
\end{enumerate}

\subsubsection{Aggregation and Noise Injection}
The main non-granular dataset $\mathcal{D}_{agg}$ provided to the participants consists of several contingency tables, such as shown on Table\ref{tab:sample_granular_data}, providing counts of displays, clicks and sales from $\mathcal{D}_{raw}$.
Each pair of features corresponds to one Table. Each of those $19 \times 18 / 2 = 171$ tables is the result of a SLQ query such as:
\begin{verbatim}
SELECT feat1, feat2, COUNT, SUM(click), SUM(sales)
FROM private_dataset GROUP BY (feat1, feat2)
\end{verbatim}
 In other words, one has access to the count of examples and labels for each level 2 combination of feature modalities (a.k.a. cross-features).
We also provided, for each feature, one Table aggregating on this feature only, for a total of $171+19 = 190$ tables.

\paragraph{Noise Injection}
To obtain differential privacy \cite{dwork2006our} guarantees, we added some iid Gaussian noise of standard deviation $17$ to all counts of displays, clicks and sales in the aggregation table. The interested reader can refer to the Appendix for the details on the choice of this parameter. Finally, please note that noisy data can sometimes produce counter-intuitive statistics with, for instance, negative click-through rates.

\paragraph{Filtering data before adding the noise}
When building $\mathcal{D}_{agg}$, the pairs of modalities that had a count smaller than 10 have been discarded to keep the dataset small. The true count (i.e., before the addition of the Gaussian noise) was used, and this is not in theory compatible with differential privacy. The "correct" methodology here would be to add noise to every pair of modalities (even those with a count of 0) and to threshold afterward. However, since some features have up to $10^5$ modalities, the number of possible pairs of modalities would be about $10^{10}$, which would make the dataset too extensive and too challenging to use. Note that the "Hashing trick" could provide a scalable and reasonably simple solution to obtain true differential privacy on our specific setting. Details on the method are given in Appendix. 

\subsubsection{Individual-level side information}
In a real, differentially private system, individual, noise-free data such as  $\mathcal{D}_{train}$ and  $\mathcal{D}_{test}$ may not be available. They were provided in the challenge for the sake of model debugging and development. However, $\mathcal{D}_{train}$ was made small enough so that challengers could not solely rely on it. 


\subsubsection{Release of the dataset}
This paper publicly releases the whole dataset used in this challenge. 
All the {data used to run the challenge}\footnote{ \url{http://go.criteo.net/criteo-ppml-challenge-adkdd21-dataset.zip}} are provided, including   $\mathcal{D}_{train}$, $\mathcal{D}_{agg}$ and $\mathcal{D}_{test}$ with the corresponding labels.

Additionally, the {full granular dataset} $\mathcal{D}_{raw}$ with 88M lines from which the aggregated dataset $\mathcal{D}_{agg}$ was produced is also available\footnote{\url{http://go.criteo.net/criteo-ppml-challenge-adkdd21-dataset-raw-granular-data.csv.gz}}. An additional set of about 4M lines, not used in the challenge, is also provided to allow computation of more accurate validation metrics.
These raw data can be used to develop and test different aggregation strategies. 

\subsection{Notations and formal description of the aggregated dataset}
We propose a more formal definition of the aggregation process. The introduced notations will be helpful when describing the best-performing submissions.

Each line of the dataset $\mathcal{D}_{raw}$ is defined by a tuple $(x, y_c, y_s)$ where $x $ denotes the feature vector of dimension $19$, and $y_c$ and $y_s$ the binary labels corresponding to respectively clicks and sales.

\begin{figure}[t]
    \centering
    \includegraphics[scale=.5]{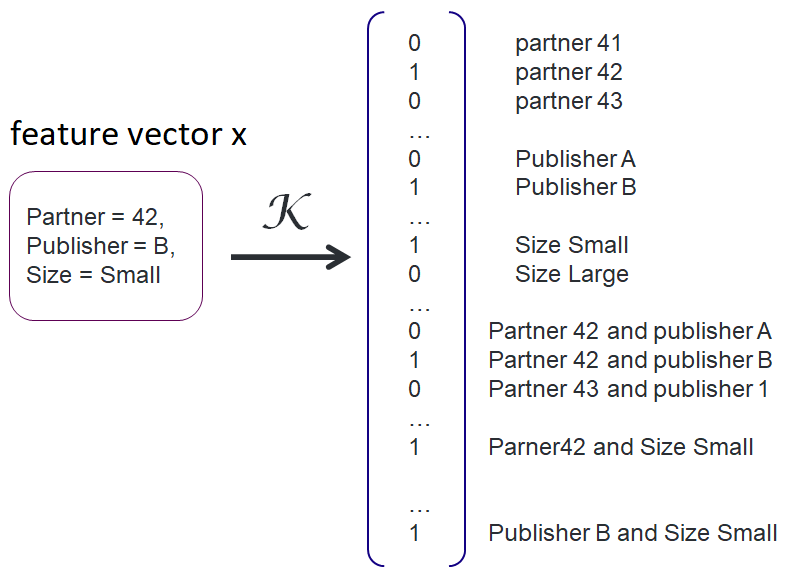}
    \caption{Illustration of the encoded vector $\mathcal{K}(x)$ with three single features: Partner, Publisher and campaign size.}
\label{fig:k_x}
\end{figure}

For $i \in 1 \hdots 19$, let $ K_i(x) \in \{0,1\}^{d_i}$ be a one hot encoding of this $i$-th feature, where we noted $d_i$ the number of distinct modalities of the $i$-th feature. We also define, for a pair $(i,j)$,  $i < j \in 1 \hdots 19$ of features, $K_{ij}(x) \in \{0,1\}^{d_i \times d_j}$ a one-hot-encoding of this pair. Formally, it may be defined as the matrix $K_{i}(x) \times K_{j}(x)^ \intercal \in \mathcal{M}_{d_i \times d_j}(\{0,1\}) $, which we view as a vector in $\{0,1\}^{d_i \times d_j}$. Finally, we define the binary vector $K(x)$ as the concatenation of all the $K_i(x)$ and $K_{i,j}(x)$ vectors:
$$K(x) := \bigoplus\limits_{i} K_i(x) \oplus  \bigoplus\limits_{i < j} K_{ij}(x) \in \{0,1\}^{ \sum_i d_i + \sum_{i<j} d_i \times d_j } $$
We propose an example of this construction in Figure \ref{fig:k_x}. 

By design, there exists a natural one-to-one mapping between the components of the vector $K(x)$ and the lines of the aggregation tables of our aggregated dataset, and an example is counted on one line of a table if and only if the associated component of $K(x)$ is 1. 

We can define the noiseless aggregated counts of displays $D$:
\begin{equation}\label{display}
    D := \sum_{(x_i, y_i) \in \mathcal{D}_{raw}}  K(x_i)
\end{equation}
Similarly, the aggregated number of clicks and sales are:
\begin{equation}\label{clicks}
    C := \sum_{(x_i, y_i^c) \in \mathcal{D}_{raw}}  y_i^c \cdot K(x_i) \quad \text{ } \quad
    S := \sum_{(x_i, y_i^s) \in \mathcal{D}_{raw}}  y_i^s \cdot K(x_i)
\end{equation}

Finally, Gaussian noise is added in order to make the data differentially private. Let $g_D , g_C, g_S$ some vectors of iid Gaussian noise of mean $0$ and variance $17^2$. The final aggregated dataset $\mathcal{D}_{agg}$ is made of the three vectors $D + g_D , C + g_C, S + g_S  $.

\subsection{Evaluation}
In the competition, participants had access to $\mathcal{D}_{test}$ which is non-aggregated, and had to predict clicks and sales labels.

These predictions were evaluated using the log-loss (a.k.a. binary cross-entropy) metric for which lower is better:
\begin{equation*}
    L = - \sum_{(x_i, y_i) \in \mathcal{D}_{test}} (y_i \log (p_i) + (1 - y_i) \log (1 - p_i))    
\end{equation*}
where $p_i$ is a prediction for $x_i$ and $y_i$ the true label. The log loss is a metric correlated to application performance for advertising, but it typically evolves on a tiny scale and is not comparable across datasets or labels.
To make the scores comparable between click and sales prediction tasks, we also report the Normalized Cross-Entropy (\nllh)  for which higher is better:
\begin{equation*}
    \nllh = (H(Y) - L) / H(Y)
\end{equation*}
where $H(Y)$ is the Shannon entropy of the label. \nllh is in $[-\infty,1]$  and can be interpreted as the relative improvement in log-loss over a model always predicting the mean label (a.k.a. "Dummy" model, which has a \nllh of $0$). It may also be viewed as the portion of label entropy explained by the model.

Finally, one could be willing to evaluate the performance of the different methods with respect to several baselines. To do so, we additionally report the degradation in log-loss over a \textit{Skyline}\footnote{it was misnamed "Oracle" on the challenge website} model. This model is learned with full access to all instances of the granular training set (without any aggregation or noise). The degradation, therefore, stresses the cost of moving from a granular, labeled dataset to an aggregated, noisy one. The Skyline model ought to be an upper bound on the performance reached by any model learned on the aggregated, noisy data.

Finally, even though models are learned on aggregated, differentially private data, the chosen evaluation metric is the accurate, noise-free indicator of the model performance. Remark that the complete access to $\mathcal{D}_{test}$ enables the use of semi-supervised techniques, which would be impossible in a real setting. 

\section{Results Summary}

\begin{figure*}[ht!]
    \centering
    \begin{minipage}[c]{0.495\linewidth}
    \includegraphics[width=\linewidth]{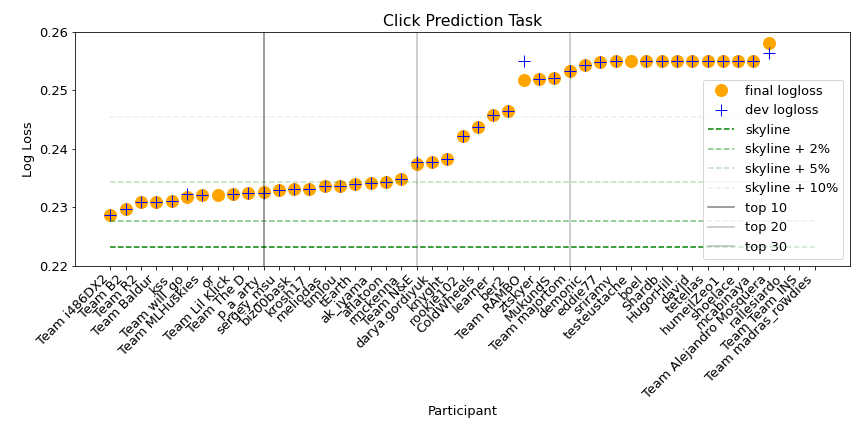}
    \end{minipage}
    \begin{minipage}[c]{0.495\linewidth}
    \includegraphics[width=\linewidth]{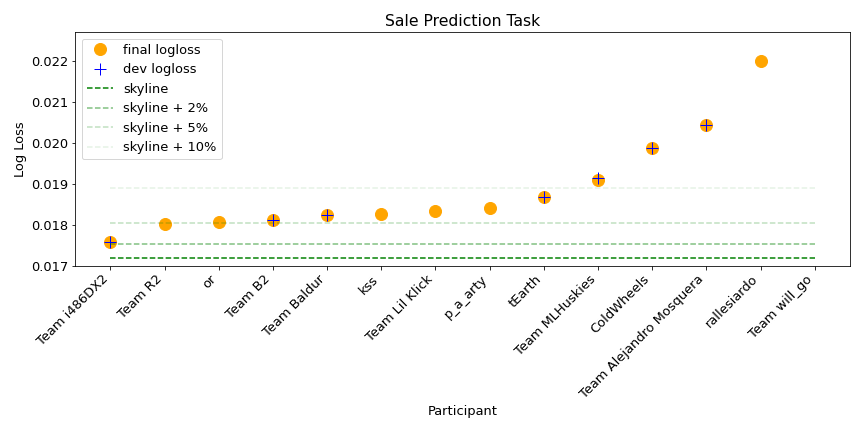}
    \end{minipage}
    \caption{Log-loss distribution among participants in Click (left) and Sale (right) prediction tasks in dev. and final phases}
    \label{fig:perf_dist}
\end{figure*}

\paragraph{Participation} The challenge attracted 177 participants over a period of 3 months (May to August 2021), of which 62 produced solutions beating a "dummy" model  - that always predicts the average label - and thus entering the final leaderboard.
Participants were free to form teams at any moment of the competition.

\paragraph{Global performance} Fig. \ref{fig:perf_dist} depicts the best performance attained by each participant or team during the development and final phases for the click and sale prediction tasks. A first remark is that the number of non-trivial solutions was higher in the click prediction task, probably because the prize money was also higher. A second remark is that there are roughly 3 groups: very good performers (20 participants) with a log-loss of up to $.235$ (that is between +2\% and +5\% vs the skyline), intermediate (7 participants) with a log-loss between $.235$ and $.250$ (+5\% to +10\% vs the skyline) and finally basic (17 participants) with a log-loss higher than $.250$ (>+10\% vs skyline). 

\paragraph{Winners performance} Table \ref{tab:click_top3} and \ref{tab:sale_top3} describe precisely the performance of the top 3 solutions for click and prediction tasks, respectively. Interestingly, the same team (i486DX2) won both with a very similar performance in terms of \nllh.
Team R2 also appears on both podiums, indicating it also found a method applicable to both tasks. We recall that both tasks are similar in data but use a different label, sale prediction being notably harder due to more randomness and less positive signal in the outcome. After performing a statistical test on the difference between each entry performance in the podium (on 10,000 bootstraps), we found the difference between the winner and runner-ups to be significant at a level of 5\%. 

\begin{table*}[h]
\begin{subtable}[t]{.45\linewidth}
\centering
\caption{Click Prediction Task}
    \begin{tabular}{l|l|l|l|l}
         Position & Participant & Log-loss  & \nllh &  Skyline\footnote{w.r.t. log loss}\\
         \hline
         Skyline & & .223184 & .312 & \\
         1st & Team i486DX2 & .228707 & .295 & -2.47\% \\
         2nd & Team B2 & .229662 & .292 & -2.90\% \\
         3rd & Team R2 & .230858 & .288 & -3.44\% \\
         Dummy & & .324474 & .0 & \\
    \end{tabular}
\label{tab:click_top3}%
\end{subtable}%
\quad
\begin{subtable}[t]{.45\linewidth}
\centering
\caption{Sale Prediction Task}
    \begin{tabular}{l|l|l|l|l}
         Position & Participant & Log-loss & \nllh & vs Skyline\footnote{w.r.t. log loss} \\
         \hline
         Skyline & & .017179 & .338 & \\
         1st & Team i486DX2 & .017583 & .322 & -2.35\% \\
         2nd & Team R2 & .018008 & .306 & -4.82\% \\
         3rd & or & .018074 & .303 & -5.21\% \\
         Dummy & & .025945 & .0 & \\
    \end{tabular}
\label{tab:sale_top3}%
\end{subtable}
\caption{Performances of the winning teams on both tasks: click and sale prediction. The take-home message is that, under the proposed setting, the loss in performance of the winning method remained limited to less than $2.5\%$ compared to the Skyline.}
\end{table*}

\subsection{The winning solutions}
\label{sec:winning_solutions}
Overall, two distinct families of methods have performed well on the challenge:
\begin{enumerate}
    \item a first \textit{Enriching method} consisted in enriching the small train set with features computed from aggregated data (a.k.a. Target Encoding \cite{pargent2021regularized}),
    \item the second one, \textit{Aggregated Logistic}, trained a logistic model with aggregated labels and unlabeled granular data.
\end{enumerate}
The first method was the one used by most teams and allowed to reach the Top 3. Only the top 2 challengers used the second method or a variant. 
Also,  most teams used ensembling techniques such as bagging. 
These techniques are widely used in a challenge, but only provide incremental improvements, 
and we did not investigate those methods further in this paper. 
Videos from the top-3 teams describing the best solutions are available online\footnote{ \url{https://go.criteo.net/AdKDD21vid}}; for conciseness we formalize them now.

\subsubsection{Enriching method} A simple but successful method consists in augmenting the training set $\mathcal{D}_{train}$ with features computed from the aggregated dataset $\mathcal{D}_{agg}$. By doing so, one can later apply classical ML methods to this enriched dataset.

To define these additional features, one can compute for each feature the observed CTRs in $\mathcal{D}_{agg}$. 
Let $x \in \mathcal{D}_{train}$. For each pair of (feature, modality), corresponding to the $i$th coordinate in $\mathbb{M}$, we can, using \eqref{display} and \eqref{clicks}, add a new feature defined by $C_{i} / D_{i}$ , where $C_{i}$ and $D_{i}$ are the counts of clicks and displays on this specific modality. Doing so, we may thus define 19 additional columns. Since we also have access to data aggregated by pair of features, we may similarly define one CTR on each pair of features; and add those CTRs as $19\times18\times0.5 = 172 $ additional columns (one per pair). Another possibility is also to add columns for the count of displays on each (pair of) modality.
Once those enriched features are defined, any classical ML model may be fitted on the small train with rich features.

In our preliminary experiments,  GBDT \cite{chen2016xgboost} has been outperforming both logistic regressions and neural nets in this setting, and we thus retained it in the presented experiments.
We also regularized the rich CTRs with a beta prior to alleviating the issues caused by the noise.
There are many different ways to tune and improve this method, either by working on the rich feature definitions or the ML models and their meta parameters. In our experiments, we kept it reasonably simple, explaining why the top challenger submissions are marginally better than our implementation. We refer to the videos of the challengers for more details.


\subsubsection{Aggregated logistic}\label{sec:agglogistic}
The main idea of this method is that having access to aggregated labels along with granular unlabeled data is sufficient to train a logistic model.

\paragraph{Notations} Let $(x, y) \in \mathcal{D}_{raw}$, and $K(x)$ be the binary encoding of the vector $x$. We note $\sigma$ the sigmoid function, and $\theta$ a parameter vectors, of the same size as $K(x)$. Finally, $D$ and $C$ are the aggregated displays and label sum defined in \eqref{display} and \eqref{clicks}. 

\paragraph{Logistic model}
We propose to model the probability of a click using a logistic regression:
\begin{equation}\label{eq:logistic}
 P_\theta( x ) := \hat P_\theta( Y=1|X=x ) := \sigma( \theta \cdot K(x)),
\end{equation}
Where $\theta \in \Theta$ refers to the classifier's parameters and $\sigma$ to the sigmoid function. Note that this corresponds to a logistic model with all cross-features. Thus, the following lemma is now a standard result:
\begin{lemma}
The gradient of the log-likelihood of the logistic model on the (unobserved) dataset $\mathcal{D}_{raw}$ is:
\begin{equation}\label{eq:gradient}
     \nabla_\theta LLH  =  \sum\limits_{(x,y) \in \mathcal{D}_{raw}} y \cdot K(x)  -  \sum\limits_{(x,y) \in \mathcal{D}_{raw}} P_\theta( x ) \cdot K(x) 
\end{equation}
\end{lemma}
In line with \eqref{clicks}, the first term $\sum\limits_{ x,y \in \mathcal{D}_{raw} }y \cdot K(x)$ may be approximated (if we ignore the Gaussian noise) by the noisy aggregated click vector $C$. On the other hand, the second term $\sum\limits_x P_\theta( x ) \cdot K(x)$, can be interpreted as the aggregated predictions of the model on $\mathcal{D}_{raw}$. Since the predictions need to be inferred on granular data, this term cannot be computed yet. To do so, let assume that we have another set $\mathcal{D}_{granular}$ of (unlabeled) granular samples of the same distribution. 

We can use those samples to estimate the second term of equation \eqref{eq:gradient}, by computing the mean of $P_\theta( x ) \cdot K(x)$ on this specific set $\mathcal{D}_{granular}$, and rescaling to the number of samples in $\mathcal{D}_{raw}$.
In other words, we can approximate the gradient of the likelihood by the following formula and apply gradient ascent. 
\begin{equation}\label{eq:gradientestimator}
\nabla_\theta LLH  \; \hat{=} \;
 C  - \frac{ \#\mathcal{D}_{raw}  }{\# \mathcal{D}_{granular}}\sum\limits_{x \in \mathcal{D}_{granular}} P_\theta( x ) \cdot K(x),
\end{equation}
where $C$ refers to the click aggregated vector defined in \eqref{clicks}
The winning team reported using ADAM for optimization purposes, while we used a preconditioned batch gradient with constant step size in our own experiments.

\paragraph{Regularization}
As is standard when learning a logistic model, a regularization may be added and significantly improves the generalization of the trained model. The winning team reported using both an L1 and L2 penalty; we used only L2 in our experiments.

\paragraph{Coordinate-wise rescaling}
In equation \ref{eq:gradientestimator}, the factor $\frac{ \#\mathcal{D}_{raw}  }{\# \mathcal{D}_{granular}}$ rescales globally by the relative sizes of the datasets. It is possible to improve this by rescaling each coordinate by the ratio of counts of samples observed in each dataset on this coordinate. The gradient is then estimated by the following equation where multiplication $\odot$ and division are coordinate-wise.

\begin{equation}
\label{eq:rescaledgradient}
\nabla_\theta LLH  \; \hat{=} \;
 C  -
\frac{ \sum\limits_{ x \in \mathcal{D}_{raw} } K(x)  } { \sum\limits_{ x \in \mathcal{D}_{granular} } K(x)  } \odot
\sum\limits_{ x \in \mathcal{D}_{granular} }   P_\theta( x ) \cdot K(x)  
\end{equation}
Here also, we may approximate $\sum\limits_{ x \in \mathcal{D}_{raw} } K(x)$ by the noisy aggregated displays $D$ (the only difference is that $D$ contains some centered Gaussian noise which we ignore here). 

\paragraph{Using test samples $\mathcal{D}_{test}$}
In the challenge, the largest available set of granular samples was $\mathcal{D}_{test}$. The challenge's winner thus used those samples as $\mathcal{D}_{granular}$ in equation \ref{eq:rescaledgradient}.

\section{Additional Experiments}
In this section, we want to understand and compare approaches for the challenge's tasks. The three methods previously presented are the \textit{Enriching method}, the \textit{Aggregated logistic}, and the \textit{Skyline} (with access to a larger labeled granular data).
More specifically, we target the following questions: 
\begin{enumerate}
    \item What is the impact of the privacy noise level on both the \textit{Enriching method} and \textit{Aggregated Logistic}? 
    \item What is the impact of the number of granular samples on these three methods?
    \item How to adapt the \textit{Aggregated logistic} to aggregated data only, and what is the resulting loss in terms of performance?  
    \item On the winning \textit{Aggregated logistic} method, What is the importance of the rescaling, and of the access to test samples  $X_{test}$ at training time?
\end{enumerate}

\paragraph{Reproducibility}
All the code to reproduce the experiments described here will be made available on Github. 
We made several runs of each experiment containing some random element. The observed variance between those runs was negligible, typically less than $10^{-4}$, and we thus omitted to report the confidence intervals.
The reported metric itself is sensitive to the set of samples on which it is computed. However, we observed that the difference between models for this metric is very stable, and always report the metric on $\mathcal{D}_{test}$.
For example, bootstrap's statistical tests confirmed that the top 3 submissions in the challenge were significant at a confidence level of 5\%, even on the noisier sales task.

\subsection{Noise level}
\begin{figure*}
\begin{subfigure}{.45\textwidth}
    \centering
    \includegraphics[width=.75\linewidth]{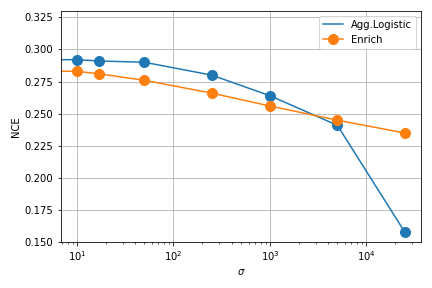}
    \caption{Click Task}
\end{subfigure}%
\quad
\begin{subfigure}{.45\textwidth}
    \centering
    \includegraphics[width=.75\linewidth]{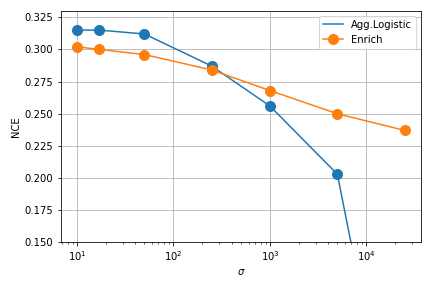}
    \caption{Sales Task}
\end{subfigure}
    \caption{Performance when varying privacy noise level $\sigma$ \textmd{-- noise affects the learning mildly up to a point where AggLogistic fails. The cross point for sales occurs earlier since by nature the counts of sales are lower and thus more sensitive to the additional noise.}}
\label{fig:varying_noise_level}
\end{figure*}

In the challenge, a noise with a standard deviation of 17 was added to the data. Here we trained both the \textit{Enriching method} and the \textit{Aggregated Logistic} on a wide variety of noise levels and compared the results.
Note that the release of the granular data enables to recompute the noiseless aggregated data, and then test the performances when the noise level is set lower than during the competition.

We trained each method with a number of granular samples equal to what was available in the challenge which means about $1e5$ samples for \textit{Enrich} and $1e6$ for \textit{Aggregated logistic}.

Figure \ref{fig:varying_noise_level} reports the results for the click and sale prediction task. For the \textit{Aggregated logistic}, we benched the L2 regularization parameter on a $\log_4$ scale and reported the best result only. For \textit{Enrich} we benched the prior weight for CTR regularization.

We observe here that the two methods are not equally impacted by the noise:
Enrich method seems quite resilient to a high level of noise, with a linear decrease in performance as a function of it. Remark that the no-noise case is not on the plot since its representation in log scale is problematic but we observe an immediate drop of performance since the values are resp. $0.289$ and $0.311$ for resp. clicks and sales for \nllh. 

At reasonable noise level the impact on  AggLogistic is small. In particular, 
the difference of performances of this method between the challenge setting ( $\sigma=17$ ) and the noiseless case is almost negligible (less than $1e-3$ in $\nllh$). We therefore believe that the main difficulty in the regime of this competition was dealing with the aggregated data, not especially with the noise. By contrast, at higher noise level, the performance drops below the \textit{Enrich} strategy suggesting that it may be possible to improve the algorithm by incorporating a modelization of the noise.

\subsection{Granular data size}

\begin{figure*}
\begin{subfigure}{.45\textwidth}
    \centering
    \includegraphics[width=.75\linewidth]{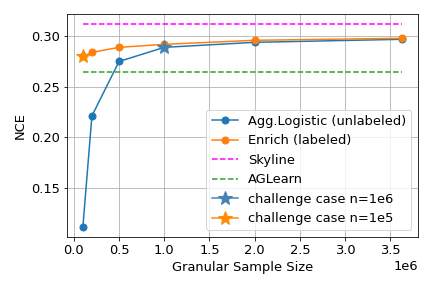}
    \caption{Methods dependence on available granular data \textmd{-- AggLogistic is better than Enrich with plenty of unlabelled, granular data}} \label{fig:varying_granular_size}
\end{subfigure}
\quad
\begin{subfigure}{.45\textwidth}
    \centering
    \includegraphics[width=.75\linewidth]{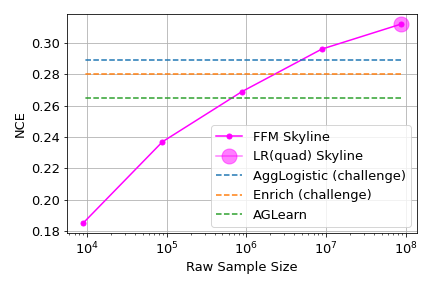}
    \caption{Skyline performance with Data Size \textmd{-- studied privacy setups degrade NCE as if data was sampled by 1 to 2 orders of magnitude}} \label{fig:skyline_perf_data_size}
\end{subfigure}
    \caption{Impact of the size of the dataset on the different methods - Click Task}
\end{figure*}

\paragraph{Unlabeled data}

In this set of experiments, we varied the size of the granular data sets used by either \textit{Enriching} or \textit{Aggregated Logistic} method.
We used here the $3.6e6$ kept out samples released with the full dataset to increase this size above what was available in the competition.
In both cases, the aggregated dataset $\mathcal{D}_{agg}$ is kept fixed ,with $\sigma = 17$.
As in the previous experiment, we crossvalidated the main parameters on both methods and reported the best setting in Figure \ref{fig:varying_granular_size}.
When the number of granular samples becomes high enough (over $1.5e6$), both methods seem equivalent in terms of performance, but the \textit{Aggregated Logistic} does not need the labels.
\paragraph{Labeled data for the Skyline}
In Figure \ref{fig:skyline_perf_data_size}, we analyze the impact of reducing the data size on the Skyline performance for two different algorithms. First, a Logistic Regression model with a quadratic feature kernel and a hashing trick \textit{LR (quad)} \cite{chapelle2014simple}. Second, we benchmarked a Field-aware Factorization Machine \textit{FFM} with 4 latent factors \cite{juan2016field} as implemented in LibFFM. 

For low sample sizes, both the \textit{Aggregated logistic} and the \textit{Enriching} method outperform the Skyline. This is due to the fact that these former methods have always access to $\mathcal{D}_{agg}$ computed from 88M samples which give them an edge over the Skyline. Another insight is that when using all the available granular data (3.5M samples) along with $\mathcal{D}_{agg}$, the performance of the \textit{Aggregated logistic} winning solution is approximately equivalent to the Skyline learned on 10\% of the dataset $\mathcal{D}_{raw}$. This illustrates the strong impact of losing all granular labeled data.

\subsection{Aggregated data only}
The Aggregated logistic method is closely related to the proposal and code from \citet{gilotte2021learning}
who trains a logistic model using only the aggregated data. It could be coarsely described as:
\begin{enumerate}
    \item fit a generative parametric model on the distribution of samples using the aggregated counts,
    \item sample a fake granular set $\mathcal{D}_{fake}$ from this generative model,
    \item use this set $\mathcal{D}_{fake}$ of ``fake'' samples to estimate the gradient of the likelihood with \eqref{eq:gradientestimator} or \eqref{eq:rescaledgradient}.
\end{enumerate}
This method obtains a NCE of 0.265 on the click competition and 0.284 on the sales competition at  a higher computational cost.

\section{Ablation studies}
\label{apx:ablation}

\begin{figure}[h]
    \centering
    \includegraphics[width=.75\linewidth]{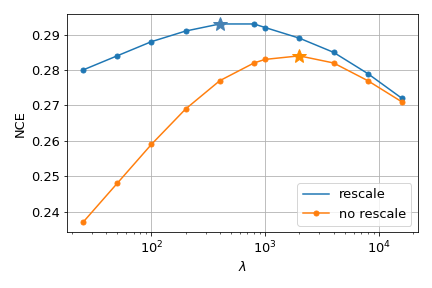}
    \caption{Ablation studies (Click Task) - Rescaling trick improves performance \textmd{-- even more when the training model is less regularized.} \label{fig:ablation_rescaling}}
\end{figure}

\paragraph{Usefulness of the re-scaling trick}
To better understand the efficiency of the re-scaling trick, Figure \ref{fig:ablation_rescaling} compares the results of the \textit{Aggregated logistic} when using either the "simple" gradient of \eqref{eq:gradient} with the "rescaling trick" defined in \eqref{eq:rescaledgradient}, as a function of the L2 regularization parameter. This rescaling always brings an improvement, all the more for lower values of noise.

\begin{figure}[h]
\label{fig:ablation_test}
    \centering
    \includegraphics[width=.75\linewidth]{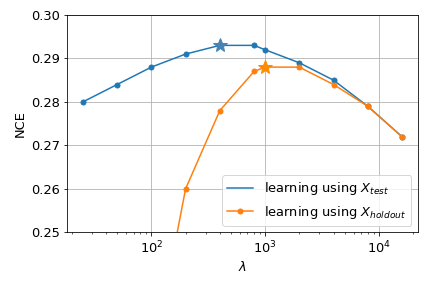}
    \caption{Ablation studies (Click Task) - Access to exact test samples also explains performance}.
\label{fig:ablation}
\end{figure}

\paragraph{Regularization when using unlabeled granular (test) samples}
In the challenge, the winners used the granular data of the \textit{test} set for training an AggLogistic. In most applications however, it is not be reasonable to require having access to the test samples to train a model. We thus compared the performances when training with these test samples, to the performances when training with another set of i.i.d. samples. The results are displayed in Figure \ref{fig:ablation}, as a function of the regularisation strength. We may first note that having access to the test samples does bring an increase. This increase is however not game-changing if the regularization is correctly returned. We may also notice that when the regularisation is lowered, the performances of the model trained on other samples drop sharply while it is not the case when using test samples. It seems that with a low regularisation, we overfit the \textit{unlabeled samples} used for training. Note this happens without observing the labels of those samples! We checked carefully its existence and lack a clear explanation for it.  
We conjecture that it is due to the combination of several passes of gradient descent on rare combinations of features.

\section{Conclusion}


\begin{table}[h!]
\resizebox{.95\linewidth}{!}{%
    \centering
    \begin{tabular}{l|c|cc|cc|c}
& \multicolumn{6}{c}{Available Granular Data} \\
 & Full: & \multicolumn{2}{c}{Plenty:} & \multicolumn{2}{c}{Scarce:} & None: \\
Privacy & 100\% & \multicolumn{2}{c}{> 1\%} & \multicolumn{2}{c}{<= 0.1\%} & 0\% \\
Noise ($\sigma)$ & & Lab & Un & Lab & Un & \\
    \hline
$\sigma=0$ & FFM \good{\Smiley} & EM \good{\Smiley} & \underline{AL} \good{\Smiley} & \underline{EM} \good{\Smiley} & AL \neutral{\Neutrey} & \cite{gilotte2021learning} \neutral{\Neutrey} \\
$\sigma<10^3$ & & EM \good{\Smiley} & \underline{AL} \good{\Smiley} & \underline{EM} \good{\Smiley} & AL \neutral{\Neutrey} & \cite{gilotte2021learning} \neutral{\Neutrey} \\
$\sigma>10^3$ & & \underline{EM} \good{\Smiley} & AL \good{\Smiley} & \underline{EM} \good{\Smiley} & AL \neutral{\Neutrey} & \cite{gilotte2021learning} \neutral{\Neutrey}\\
    \end{tabular}}
    \caption{
    \textmd{Findings- EM: \textit{Enriching method}, AL: \textit{Aggregated Logistic}.
     \label{tab:summary}}}
\end{table}

\paragraph{Findings}
In Table \ref{tab:summary} we summarize what we learned with the challenge and the additional experiments. 
The biggest surprise was how helpful is the availability of a small set of granular samples when trying to extract information from aggregated data. At second thought, it naturally finds an explanation in the explicit formula of a gradient update. 
However, the studied privacy setup incurs a loss in prediction performance comparable to sampling the dataset by 1 to 2 orders of magnitude, depending on the availability of granular data. This is significant for the Online Advertising business performance and also impacts ad relevancy and user experience. This calls for further refinement and discussions of the original proposal at the W3C. 
Finally, when no granular data is available the \textit{Aggregated Logistic} method can be considered as a fallback, yet further research is needed to put it on par with other options.

\paragraph{Limitations}
While every reasonable effort was made to anchor the challenge in reality by mimicking the mechanics proposed to the Improving Web Advertising forum at W3C, it is beyond the scope of such a challenge to simulate a live system of such complexity as online advertising.
Among the problems that we know are perhaps harder to handle in the real problem, we can list: the noise distribution, delays in data pipelines, the scale of the global system, the translation from log-loss/NCE to a business metric. In particular, we left out the question of finding optimal aggregations. 
Indeed, in the real world, it would be possible to ask different statistics than second-order ones, and the number of requests may be limited. 
While, in a real setting, it is reasonable to expect access to some granular data (e.g., "first-party" data) it is unclear that it would have the same distribution as the aggregated one, further studies on distribution shift or transfer learning would be of interest. 
Because of those real-world complications, we believe that the problem proposed in the challenge is only a reasonable upper bound on the performance that could be attained in a production system: we envision that the performance in practice would possibly be lower.\\
Finally, the performance metric used in this challenge is not a business revenue metric. Still, it should be reasonably correlated to it, at least in the case of performance-based advertising, where the payoff is linked to prediction quality through bidding.

\paragraph{Future works}
The role of the presence of a small granular dataset calls for further investigation. First, it should be possible to obtain a new convergence bound depending both on the number of granular samples and the number of aggregated ones. Secondly, the mechanism could be used to save computational costs since the aggregated data can be more than a magnitude smaller than their granular counterpart. Finally, granular samples suffering from a distribution require investigation. These methods might also find direct usage in different domains: for example, it is common for a hospital to have a relatively small set of data on their patients, which they cannot share for privacy concerns. They might share some noisy aggregated data, allowing each hospital to leverage its own small granular data, allowing a new kind of federated learning.  

\begin{acks}
We would like to acknowledge support for running this competition by the Criteo AI Lab and the AdKDD workshop organization team for providing a scientific discussion ground for the challenge.
\end{acks}

\clearpage
\bibliographystyle{ACM-Reference-Format}
\bibliography{paper}


\begin{thebibliography}{29}


\ifx \showCODEN    \undefined \def \showCODEN     #1{\unskip}     \fi
\ifx \showDOI      \undefined \def \showDOI       #1{#1}\fi
\ifx \showISBNx    \undefined \def \showISBNx     #1{\unskip}     \fi
\ifx \showISBNxiii \undefined \def \showISBNxiii  #1{\unskip}     \fi
\ifx \showISSN     \undefined \def \showISSN      #1{\unskip}     \fi
\ifx \showLCCN     \undefined \def \showLCCN      #1{\unskip}     \fi
\ifx \shownote     \undefined \def \shownote      #1{#1}          \fi
\ifx \showarticletitle \undefined \def \showarticletitle #1{#1}   \fi
\ifx \showURL      \undefined \def \showURL       {\relax}        \fi
\providecommand\bibfield[2]{#2}
\providecommand\bibinfo[2]{#2}
\providecommand\natexlab[1]{#1}
\providecommand\showeprint[2][]{arXiv:#2}

\bibitem[\protect\citeauthoryear{Abadi, Chu, Goodfellow, McMahan, Mironov,
  Talwar, and Zhang}{Abadi et~al\mbox{.}}{2016}]%
        {Abadi2016}
\bibfield{author}{\bibinfo{person}{Martin Abadi}, \bibinfo{person}{Andy Chu},
  \bibinfo{person}{Ian Goodfellow}, \bibinfo{person}{H.~Brendan McMahan},
  \bibinfo{person}{Ilya Mironov}, \bibinfo{person}{Kunal Talwar}, {and}
  \bibinfo{person}{Li Zhang}.} \bibinfo{year}{2016}\natexlab{}.
\newblock \showarticletitle{Deep Learning with Differential Privacy}. In
  \bibinfo{booktitle}{\emph{Proceedings of the 2016 ACM SIGSAC Conference on
  Computer and Communications Security}} (Vienna, Austria)
  \emph{(\bibinfo{series}{CCS '16})}. \bibinfo{publisher}{Association for
  Computing Machinery}, \bibinfo{address}{New York, NY, USA},
  \bibinfo{pages}{308–318}.
\newblock
\showISBNx{9781450341394}
\urldef\tempurl%
\url{https://doi.org/10.1145/2976749.2978318}
\showDOI{\tempurl}


\bibitem[\protect\citeauthoryear{Anderson}{Anderson}{2021}]%
        {maskedLARK}
\bibfield{author}{\bibinfo{person}{Erik Anderson}.}
  \bibinfo{year}{2021}\natexlab{}.
\newblock \bibinfo{title}{Masked Learning, Aggregation and Reporting worKflow
  (Masked LARK)}.
\newblock
  \bibinfo{howpublished}{\url{https://github.com/WICG/privacy-preserving-ads/blob/main/MaskedLARK.md}}.
\newblock
\newblock
\shownote{Accessed: 2021-05-01.}


\bibitem[\protect\citeauthoryear{Avazu}{Avazu}{2014}]%
        {avazu}
\bibfield{author}{\bibinfo{person}{Avazu}.} \bibinfo{year}{2014}\natexlab{}.
\newblock \bibinfo{title}{Avazu CTR Prediction Contest}.
\newblock
  \bibinfo{howpublished}{\url{https://www.kaggle.com/c/avazu-ctr-prediction}}.
\newblock
\newblock
\shownote{Accessed: 2021-05-01.}


\bibitem[\protect\citeauthoryear{Avito.ru}{Avito.ru}{2015}]%
        {avito}
\bibfield{author}{\bibinfo{person}{Avito.ru}.} \bibinfo{year}{2015}\natexlab{}.
\newblock \bibinfo{title}{Avito Context Ad Clicks}.
\newblock
  \bibinfo{howpublished}{\url{https://www.kaggle.com/c/avito-context-ad-clicks}}.
\newblock
\newblock
\shownote{Accessed: 2021-05-01.}


\bibitem[\protect\citeauthoryear{Belli, Tejani, Portman, Lung-Yut-Fong,
  Chamberlain, Xie, Lum, Hunt, Bronstein, Anelli, Kalloori, Ferwerda, and
  Shi}{Belli et~al\mbox{.}}{2021}]%
        {belli20212021}
\bibfield{author}{\bibinfo{person}{Luca Belli}, \bibinfo{person}{Alykhan
  Tejani}, \bibinfo{person}{Frank Portman}, \bibinfo{person}{Alexandre
  Lung-Yut-Fong}, \bibinfo{person}{Ben Chamberlain}, \bibinfo{person}{Yuanpu
  Xie}, \bibinfo{person}{Kristian Lum}, \bibinfo{person}{Jonathan Hunt},
  \bibinfo{person}{Michael Bronstein}, \bibinfo{person}{Vito~Walter Anelli},
  \bibinfo{person}{Saikishore Kalloori}, \bibinfo{person}{Bruce Ferwerda},
  {and} \bibinfo{person}{Wenzhe Shi}.} \bibinfo{year}{2021}\natexlab{}.
\newblock \bibinfo{title}{The 2021 RecSys Challenge Dataset: Fairness is not
  optional}.
\newblock
\newblock
\showeprint[arxiv]{2109.08245}~[cs.SI]


\bibitem[\protect\citeauthoryear{Bhowmik, Chen, Xing, and Rajan}{Bhowmik
  et~al\mbox{.}}{2019}]%
        {bhowmik2019estimagg}
\bibfield{author}{\bibinfo{person}{Avradeep Bhowmik}, \bibinfo{person}{Minmin
  Chen}, \bibinfo{person}{Zhengming Xing}, {and} \bibinfo{person}{Suju Rajan}.}
  \bibinfo{year}{2019}\natexlab{}.
\newblock \showarticletitle{Estimagg: A learning framework for groupwise
  aggregated data}. In \bibinfo{booktitle}{\emph{Proceedings of the 2019 SIAM
  International Conference on Data Mining}}. SIAM, \bibinfo{pages}{477--485}.
\newblock


\bibitem[\protect\citeauthoryear{Bhowmik, Ghosh, and Koyejo}{Bhowmik
  et~al\mbox{.}}{2016}]%
        {bhowmik2016sparse}
\bibfield{author}{\bibinfo{person}{Avradeep Bhowmik}, \bibinfo{person}{Joydeep
  Ghosh}, {and} \bibinfo{person}{Oluwasanmi Koyejo}.}
  \bibinfo{year}{2016}\natexlab{}.
\newblock \showarticletitle{Sparse parameter recovery from aggregated data}. In
  \bibinfo{booktitle}{\emph{International Conference on Machine Learning}}.
  PMLR, \bibinfo{pages}{1090--1099}.
\newblock


\bibitem[\protect\citeauthoryear{Bureau}{Bureau}{2021}]%
        {diffpriv_census}
\bibfield{author}{\bibinfo{person}{U.S.~Census Bureau}.}
  \bibinfo{year}{2021}\natexlab{}.
\newblock \bibinfo{title}{Differential Privacy for Census Data Explained}.
\newblock
  \bibinfo{howpublished}{\url{https://www.ncsl.org/research/redistricting/differential-privacy-for-census-data-explained.aspx}}.
\newblock
\newblock
\shownote{Accessed: 2021-05-01.}


\bibitem[\protect\citeauthoryear{CG}{CG}{2020}]%
        {measurementapi}
\bibfield{author}{\bibinfo{person}{Web~Incubator CG}.}
  \bibinfo{year}{2020}\natexlab{}.
\newblock \bibinfo{title}{The Conversion Measurement API}.
\newblock
  \bibinfo{howpublished}{\url{https://github.com/WICG/conversion-measurement-api}}.
\newblock
\newblock
\shownote{Accessed: 2021-05-01.}


\bibitem[\protect\citeauthoryear{Chapelle, Manavoglu, and Rosales}{Chapelle
  et~al\mbox{.}}{2014}]%
        {chapelle2014simple}
\bibfield{author}{\bibinfo{person}{Olivier Chapelle}, \bibinfo{person}{Eren
  Manavoglu}, {and} \bibinfo{person}{Romer Rosales}.}
  \bibinfo{year}{2014}\natexlab{}.
\newblock \showarticletitle{Simple and scalable response prediction for display
  advertising}.
\newblock \bibinfo{journal}{\emph{ACM Transactions on Intelligent Systems and
  Technology (TIST)}} \bibinfo{volume}{5}, \bibinfo{number}{4}
  (\bibinfo{year}{2014}), \bibinfo{pages}{1--34}.
\newblock


\bibitem[\protect\citeauthoryear{Chaudhuri, Monteleoni, and Sarwate}{Chaudhuri
  et~al\mbox{.}}{2011}]%
        {Chaudhuri2011}
\bibfield{author}{\bibinfo{person}{Kamalika Chaudhuri}, \bibinfo{person}{Claire
  Monteleoni}, {and} \bibinfo{person}{Anand~D. Sarwate}.}
  \bibinfo{year}{2011}\natexlab{}.
\newblock \showarticletitle{Differentially Private Empirical Risk
  Minimization}.
\newblock \bibinfo{journal}{\emph{J. Mach. Learn. Res.}} \bibinfo{volume}{12},
  \bibinfo{number}{null} (\bibinfo{date}{July} \bibinfo{year}{2011}),
  \bibinfo{pages}{1069–1109}.
\newblock
\showISSN{1532-4435}


\bibitem[\protect\citeauthoryear{Chen and Guestrin}{Chen and Guestrin}{2016}]%
        {chen2016xgboost}
\bibfield{author}{\bibinfo{person}{Tianqi Chen} {and} \bibinfo{person}{Carlos
  Guestrin}.} \bibinfo{year}{2016}\natexlab{}.
\newblock \showarticletitle{Xgboost: A scalable tree boosting system}. In
  \bibinfo{booktitle}{\emph{Proceedings of the 22nd acm sigkdd international
  conference on knowledge discovery and data mining}}.
  \bibinfo{pages}{785--794}.
\newblock


\bibitem[\protect\citeauthoryear{Criteo}{Criteo}{2014}]%
        {criteo_challenge}
\bibfield{author}{\bibinfo{person}{Criteo}.} \bibinfo{year}{2014}\natexlab{}.
\newblock \bibinfo{title}{Criteo Display Advertising Challenge}.
\newblock
  \bibinfo{howpublished}{\url{https://www.kaggle.com/c/criteo-display-ad-challenge}}.
\newblock
\newblock
\shownote{Accessed: 2021-05-01.}


\bibitem[\protect\citeauthoryear{Desfontaines}{Desfontaines}{2021}]%
        {blogdefontaines}
\bibfield{author}{\bibinfo{person}{Damien Desfontaines}.}
  \bibinfo{year}{2021}\natexlab{}.
\newblock \bibinfo{title}{The magic of Gaussian noise}.
\newblock
  \bibinfo{howpublished}{\url{https://desfontain.es/privacy/gaussian-noise.html}}.
\newblock
\newblock
\shownote{Accessed: 2021-05-01.}


\bibitem[\protect\citeauthoryear{Dwork, Kenthapadi, McSherry, Mironov, and
  Naor}{Dwork et~al\mbox{.}}{2006}]%
        {dwork2006our}
\bibfield{author}{\bibinfo{person}{Cynthia Dwork}, \bibinfo{person}{Krishnaram
  Kenthapadi}, \bibinfo{person}{Frank McSherry}, \bibinfo{person}{Ilya
  Mironov}, {and} \bibinfo{person}{Moni Naor}.}
  \bibinfo{year}{2006}\natexlab{}.
\newblock \showarticletitle{Our data, ourselves: Privacy via distributed noise
  generation}. In \bibinfo{booktitle}{\emph{Annual International Conference on
  the Theory and Applications of Cryptographic Techniques}}. Springer,
  \bibinfo{pages}{486--503}.
\newblock


\bibitem[\protect\citeauthoryear{Dwork, Roth, et~al\mbox{.}}{Dwork
  et~al\mbox{.}}{2014}]%
        {dwork2014algorithmic}
\bibfield{author}{\bibinfo{person}{Cynthia Dwork}, \bibinfo{person}{Aaron
  Roth}, {et~al\mbox{.}}} \bibinfo{year}{2014}\natexlab{}.
\newblock \showarticletitle{The algorithmic foundations of differential
  privacy.}
\newblock \bibinfo{journal}{\emph{Found. Trends Theor. Comput. Sci.}}
  \bibinfo{volume}{9}, \bibinfo{number}{3-4} (\bibinfo{year}{2014}),
  \bibinfo{pages}{211--407}.
\newblock


\bibitem[\protect\citeauthoryear{Gilotte and Rohde}{Gilotte and Rohde}{2021}]%
        {gilotte2021learning}
\bibfield{author}{\bibinfo{person}{Alexandre Gilotte} {and}
  \bibinfo{person}{David Rohde}.} \bibinfo{year}{2021}\natexlab{}.
\newblock \showarticletitle{Learning a logistic model from aggregated data}.
\newblock  (\bibinfo{year}{2021}).
\newblock


\bibitem[\protect\citeauthoryear{Harrisson}{Harrisson}{2020}]%
        {aggreportapi}
\bibfield{author}{\bibinfo{person}{C.~S. Harrisson}.}
  \bibinfo{year}{2020}\natexlab{}.
\newblock \bibinfo{title}{The Aggregate Reporting API}.
\newblock
  \bibinfo{howpublished}{\url{https://github.com/csharrison/aggregate-reporting-api}}.
\newblock
\newblock
\shownote{Accessed: 2021-05-01.}


\bibitem[\protect\citeauthoryear{Juan, Zhuang, Chin, and Lin}{Juan
  et~al\mbox{.}}{2016}]%
        {juan2016field}
\bibfield{author}{\bibinfo{person}{Yuchin Juan}, \bibinfo{person}{Yong Zhuang},
  \bibinfo{person}{Wei-Sheng Chin}, {and} \bibinfo{person}{Chih-Jen Lin}.}
  \bibinfo{year}{2016}\natexlab{}.
\newblock \showarticletitle{Field-aware factorization machines for CTR
  prediction}. In \bibinfo{booktitle}{\emph{Proceedings of the 10th ACM
  conference on recommender systems}}. \bibinfo{pages}{43--50}.
\newblock


\bibitem[\protect\citeauthoryear{King, Tanner, and Rosen}{King
  et~al\mbox{.}}{2004}]%
        {king2004ecological}
\bibfield{author}{\bibinfo{person}{Gary King}, \bibinfo{person}{Martin~A
  Tanner}, {and} \bibinfo{person}{Ori Rosen}.} \bibinfo{year}{2004}\natexlab{}.
\newblock \bibinfo{booktitle}{\emph{Ecological inference: New methodological
  strategies}}.
\newblock \bibinfo{publisher}{Cambridge University Press}.
\newblock


\bibitem[\protect\citeauthoryear{Kone{\v{c}}n{\`y}, McMahan, and
  Ramage}{Kone{\v{c}}n{\`y} et~al\mbox{.}}{2015}]%
        {konevcny2015federated}
\bibfield{author}{\bibinfo{person}{Jakub Kone{\v{c}}n{\`y}},
  \bibinfo{person}{Brendan McMahan}, {and} \bibinfo{person}{Daniel Ramage}.}
  \bibinfo{year}{2015}\natexlab{}.
\newblock \showarticletitle{Federated optimization: Distributed optimization
  beyond the datacenter}.
\newblock \bibinfo{journal}{\emph{arXiv preprint arXiv:1511.03575}}
  (\bibinfo{year}{2015}).
\newblock


\bibitem[\protect\citeauthoryear{Liao, Peng, Liu, and Shen}{Liao
  et~al\mbox{.}}{2014}]%
        {liao2014ipinyou}
\bibfield{author}{\bibinfo{person}{Hairen Liao}, \bibinfo{person}{Lingxiao
  Peng}, \bibinfo{person}{Zhenchuan Liu}, {and} \bibinfo{person}{Xuehua Shen}.}
  \bibinfo{year}{2014}\natexlab{}.
\newblock \showarticletitle{iPinYou global rtb bidding algorithm competition
  dataset}. In \bibinfo{booktitle}{\emph{Proceedings of the Eighth
  International Workshop on Data Mining for Online Advertising}}.
  \bibinfo{pages}{1--6}.
\newblock


\bibitem[\protect\citeauthoryear{McMahan, Holt, Sculley, Young, Ebner, Grady,
  Nie, Phillips, Davydov, Golovin, et~al\mbox{.}}{McMahan
  et~al\mbox{.}}{2013}]%
        {mcmahan2013ad}
\bibfield{author}{\bibinfo{person}{H~Brendan McMahan}, \bibinfo{person}{Gary
  Holt}, \bibinfo{person}{David Sculley}, \bibinfo{person}{Michael Young},
  \bibinfo{person}{Dietmar Ebner}, \bibinfo{person}{Julian Grady},
  \bibinfo{person}{Lan Nie}, \bibinfo{person}{Todd Phillips},
  \bibinfo{person}{Eugene Davydov}, \bibinfo{person}{Daniel Golovin},
  {et~al\mbox{.}}} \bibinfo{year}{2013}\natexlab{}.
\newblock \showarticletitle{Ad click prediction: a view from the trenches}. In
  \bibinfo{booktitle}{\emph{Proceedings of the 19th ACM SIGKDD international
  conference on Knowledge discovery and data mining}}.
  \bibinfo{pages}{1222--1230}.
\newblock


\bibitem[\protect\citeauthoryear{Outbrain}{Outbrain}{2016}]%
        {outbrain}
\bibfield{author}{\bibinfo{person}{Outbrain}.} \bibinfo{year}{2016}\natexlab{}.
\newblock \bibinfo{title}{Outbrain Click Prediction Challenge}.
\newblock
  \bibinfo{howpublished}{\url{https://www.kaggle.com/c/outbrain-click-prediction}}.
\newblock
\newblock
\shownote{Accessed: 2021-05-01.}


\bibitem[\protect\citeauthoryear{Papernot, Abadi, Úlfar Erlingsson,
  Goodfellow, and Talwar}{Papernot et~al\mbox{.}}{2017}]%
        {Papernot2017}
\bibfield{author}{\bibinfo{person}{Nicolas Papernot}, \bibinfo{person}{Martín
  Abadi}, \bibinfo{person}{Úlfar Erlingsson}, \bibinfo{person}{Ian
  Goodfellow}, {and} \bibinfo{person}{Kunal Talwar}.}
  \bibinfo{year}{2017}\natexlab{}.
\newblock \bibinfo{title}{Semi-supervised Knowledge Transfer for Deep Learning
  from Private Training Data}.
\newblock
\newblock
\showeprint[arxiv]{1610.05755}~[stat.ML]


\bibitem[\protect\citeauthoryear{Pargent, Pfisterer, Thomas, and
  Bischl}{Pargent et~al\mbox{.}}{2021}]%
        {pargent2021regularized}
\bibfield{author}{\bibinfo{person}{Florian Pargent}, \bibinfo{person}{Florian
  Pfisterer}, \bibinfo{person}{Janek Thomas}, {and} \bibinfo{person}{Bernd
  Bischl}.} \bibinfo{year}{2021}\natexlab{}.
\newblock \showarticletitle{Regularized target encoding outperforms traditional
  methods in supervised machine learning with high cardinality features}.
\newblock \bibinfo{journal}{\emph{arXiv preprint arXiv:2104.00629}}
  (\bibinfo{year}{2021}).
\newblock


\bibitem[\protect\citeauthoryear{Samarati and Sweeney}{Samarati and
  Sweeney}{1998}]%
        {samarati1998protecting}
\bibfield{author}{\bibinfo{person}{Pierangela Samarati} {and}
  \bibinfo{person}{Latanya Sweeney}.} \bibinfo{year}{1998}\natexlab{}.
\newblock \showarticletitle{Protecting privacy when disclosing information:
  k-anonymity and its enforcement through generalization and suppression}.
\newblock  (\bibinfo{year}{1998}).
\newblock


\bibitem[\protect\citeauthoryear{Wilander}{Wilander}{2021}]%
        {pcm}
\bibfield{author}{\bibinfo{person}{John Wilander}.}
  \bibinfo{year}{2021}\natexlab{}.
\newblock \bibinfo{title}{Private Click Measurement}.
\newblock
  \bibinfo{howpublished}{\url{https://webkit.org/blog/11529/introducing-private-click-measurement-pcm/}}.
\newblock
\newblock
\shownote{Accessed: 2021-05-01.}


\bibitem[\protect\citeauthoryear{Zhang, Charoenphakdee, Wu, and Sugiyama}{Zhang
  et~al\mbox{.}}{2020}]%
        {zhang2020learning}
\bibfield{author}{\bibinfo{person}{Yivan Zhang}, \bibinfo{person}{Nontawat
  Charoenphakdee}, \bibinfo{person}{Zhenguo Wu}, {and} \bibinfo{person}{Masashi
  Sugiyama}.} \bibinfo{year}{2020}\natexlab{}.
\newblock \showarticletitle{Learning from Aggregate Observations}.
\newblock \bibinfo{journal}{\emph{Advances in Neural Information Processing
  Systems}}  \bibinfo{volume}{33} (\bibinfo{year}{2020}).
\newblock


\end{thebibliography}

\newpage
\appendix
\section{Detailed computation of the privacy noise from a given privacy budget}
To obtain differential privacy guarantees, aggregating the data is not sufficient, it is also required to add some noise.  We opted for ($\epsilon,\delta$) differential privacy \cite[][Definition 2.4]{dwork2014algorithmic}, with a target $\epsilon=10$ and $\delta=1e-10$.
A standard way to get this privacy is to add some iid. centered Gaussian noise to each record of the aggregated data, whose variance depends on the parameters $\epsilon$ and $\delta$, and on the "L2-sensitivity" of the released data. "L2-sensitivity" measures how the addition or removal of a single record \footnote{Assuming here we want to protect the privacy of single records of the dataset. Note that if a user-contributed to several records, the required noise would be higher.} may change the L2 norm of the data. In our case, one record may change exactly one line per Table, and on those lines, it may change the counts of displays, clicks, and sales each by at most 1. With $190$ tables, this means that the L2-sensitivity is $\sqrt(190 \times 3)$ and following \cite{blogdefontaines} we obtained a standard deviation of $17$.

Note that there is a simple re-parameterization trick, namely replacing the counts of sales, clicks, and displays by the counts of sales, clicks with no sale, and displays with no click, which could simply reduce this L2-sensitivity to $\sqrt(190)$ by ensuring that a single report is counted in only one of those 3 metrics. If this re-parameterization had been used, the noise with std $17$ would have provided the $\epsilon=5$ guarantee that was mistakenly announced earlier in the challenge. 


\section{Strict Differentially Private Aggregation with the hashing trick}
The aggregated data set was not strictly differential private because we removed the pairs of modalities with a count of examples less than 10, before adding the Gaussian noise.
A simple change to the aggregation function may however avoid this issue: instead of aggregating on pairs of modalities, we may aggregate on hashes of those pairs. 
Formally, we redefine the kernel $K$ as follow:
\begin{itemize}
\item let $p$ the size of the hashing space
\item for each feature $f$, define $K_f(x)$ as a vector of size p made of 0s, with a 1 at the index $hash(f, x_f ) \pmod p$
\item similarly for each pair $f,g$ of features, define $K_{f,g}(x)$ as  vector with a 1 at index  
$hash( f,g, x_f , x_g) \pmod p$
\item finally redefine $K(x)$ as $\sum\limits_f K_f(x) + \sum\limits_{f<g} K_{f,g}(x) $
\end{itemize}
With this new definition of $K$, we may redefine the aggregated data by the equation \ref{display}, and \ref{clicks}.

Note that the $K$ kernel defined this way is exactly the "hashing trick" commonly used when learning a large-scale logistic regression with cross-features \cite{chapelle2014simple}. The equation \ref{eq:logistic} thus becomes the "usual" logistic regression with hashing trick. 
We trained the "aggregated logistic" on such hashed aggregated data, and observed only minimal degradation of the performances. With a hashing space of size 2**24, sigma=17, L2=1000 we obtained a NCE of 0.291 (instead of 0.292 with the same setting and no hashing.)

\end{document}